\documentclass[10pt,twocolumn,letterpaper]{article}

\usepackage{cvpr}              % To produce the CAMERA-READY version
\definecolor{cvprblue}{rgb}{0.21,0.49,0.74}
\usepackage[pagebackref,breaklinks,colorlinks,allcolors=cvprblue]{hyperref}
\usepackage{stfloats}
\usepackage{amsmath}
\def\paperID{329} % *** Enter the Paper ID here
\def\confName{CVPR}
\def\confYear{2025}
\usepackage{array}

\usepackage{colortbl}
\definecolor{Gray}{gray}{0.93}
\usepackage{multirow}

\title{Unleashing the Potential of Consistency Learning \\ for Detecting and Grounding Multi-Modal Media Manipulation 
}

\makeatletter
\newcommand{\myfnsymbol}[1]{%
  \expandafter\@myfnsymbol\csname c@#1\endcsname
}
\newcommand{\@myfnsymbol}[1]{%
  \ifcase #1
  \or \TextOrMath{\textasteriskcentered}{*}% 1
  \or \TextOrMath{\textdagger}{\dagger}% 2
  \fi
}
\newcommand{\equalcontributor}{\@myfnsymbol{1}}
\newcommand{\corresponding}{\@myfnsymbol{2}}
\makeatother

\author{
Yiheng Li$^{1,2,\equalcontributor}$,
Yang Yang$^{1,2,\equalcontributor}$,
Zichang Tan$^{5, \corresponding}$,
Huan Liu$^{6}$,
Weihua Chen$^{7,\corresponding}$,
Xu Zhou$^{5}$,
Zhen Lei$^{1,2,3,4}$  \\
$^1$ MAIS, Institute of Automation, Chinese Academy of Sciences\\
$^2$ School of Artificial Intelligence, University of Chinese Academy of Sciences\\
$^3$ CAIR, HKISI, Chinese Academy of Sciences \\
$^4$ School of Computer Science and Engineering, the Faculty of Innovation Engineering, M.U.S.T\\
$^5$ Sangfor Technologies Inc. \quad $^6$ Beijing Jiaotong University \quad $^7$ Alibaba Group \\
{\tt\small \{liyiheng2024, yangyang2013, zhen.lei\}@ia.ac.cn, tanzichang@foxmail.com} 
}
\begin{document}
\maketitle
\renewcommand{\thefootnote}{\myfnsymbol{footnote}}

\footnotetext[1]{Yiheng Li and Yang Yang contributed equally to this work.}%
\footnotetext[2]{Corresponding authors.}%
\setcounter{footnote}{0}% Restart footnote counter
\renewcommand{\thefootnote}{\fnsymbol{footnote}}

\begin{abstract}
To tackle the threat of fake news, the task of detecting and grounding multi-modal media manipulation (DGM$^4$) has received increasing attention. 
However, most state-of-the-art methods fail to explore the fine-grained consistency within local content, usually resulting in an inadequate perception of detailed forgery and unreliable results. In this paper, we propose a novel approach named Contextual-Semantic Consistency Learning (CSCL) to enhance the fine-grained perception ability of forgery for DGM$^4$. Two branches for image and text modalities are established, each of which contains two cascaded decoders, i.e., Contextual Consistency Decoder (CCD) and Semantic Consistency Decoder (SCD), to capture within-modality contextual consistency and across-modality semantic consistency, respectively. 
Both CCD and SCD adhere to the same criteria for capturing fine-grained forgery details. 
To be specific, each module first constructs consistency features by leveraging additional supervision from the heterogeneous information of each token pair.
Then, the forgery-aware reasoning or aggregating is adopted to deeply seek forgery cues based on the consistency features.
Extensive experiments on DGM$^4$ datasets prove that CSCL achieves new state-of-the-art performance, especially for the results of grounding manipulated content. Codes and weights are avaliable at \url{https://github.com/liyih/CSCL}.

\end{abstract}

\section{Introduction}
\label{sec:intro}

With the rapid development of the generative models \cite{ho2020denoising, goodfellow2020generative} and the large language model \cite{radford2019language}, fake media appears more frequently on the Internet \cite{juefei2022countering,zheng2020survey}, including face forgery, synthetic text, and deepfake video. This poses great threats to information security and user privacy. To solve these problems, many deepfake detection methods are proposed. Early works often focus on single-modal detection, such as face deepfake detection \cite{jeong2022bihpf, miao2023f} and text deepfake detection \cite{zhu2022generalizing, zellers2019defending}. Later works gradually focus on multi-modal data \cite{oorloff2024avff, yang2023avoid}, achieving more accurate results through the interaction between multiple modalities. 
Detecting and grounding multi-modal media manipulation (DGM$^4$) \cite{shao2023detecting} is one of the multi-modal tasks. 
% which contributes to filter the fake news in the social media.
Unlike the traditional tasks that only make binary detection (real or fake), DGM$^4$ needs to predict additional fine-grained manipulation type classification and localize the manipulated content. 

\begin{figure}[t]
  \centering
   \includegraphics[scale=0.43]{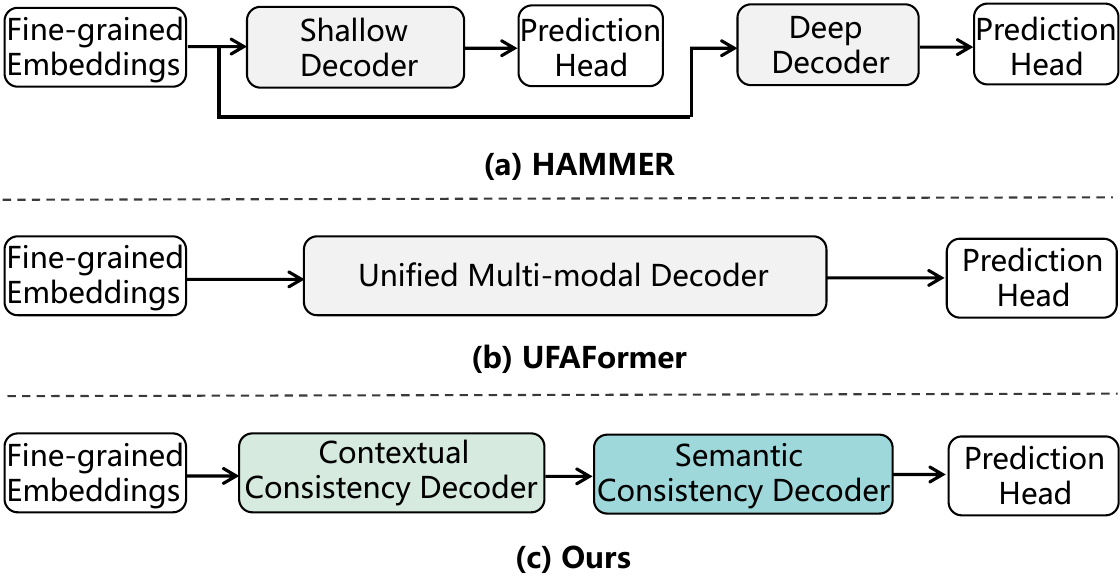}
   
   \caption{
   \textbf{Comparison of fine-grained feature process between our method and existing methods.} 
   (a) Previous methods \cite{shao2023detecting, shao2024detecting} adopt shallow and deep decoders to process embeddings for different sub-tasks.
   (b) Current SOTA methods \cite{liu2024unified, wang2024exploiting} conduct a unified multi-modal decoder for embeddings, but they ignore the consistency relationship between genuine and forged content.
   (c) Our method explores the consistency learning to achieve deeper reasoning on the DGM$^4$ task and proposes contextual and semantic consistency decoders to model the fine-grained correlation.} 

   \label{fig1}
   \vspace{-0.5cm}
\end{figure}

Many methods \cite{shao2023detecting, liu2024unified, li2024towards, wang2024exploiting, liu2024fka} are proposed for the DGM$^4$ task, 
but the results have generally been limited, particularly in the area of locating forged content.
Early methods \cite{shao2023detecting, shao2024detecting} (as shown in Fig.\ref{fig1} (a)) mainly use shallow and deep decoders to predict different kinds of sub-tasks. However, this structure limits the ability of network to learn the correlation among different sub-tasks, 
% and multi-branch structure increases the complexity of the model. 
and significant differences in the decoder structures corresponding to different sub-tasks increase the complexity of the model.
Although contrastive learning is used to establish semantic correlation across modalities \cite{he2020momentum}, the contextual consistency is ignored within a single modality.
Recent SOTA methods \cite{wang2024exploiting, liu2024unified} typically use a unified multi-modal decoder (as shown in Fig. \ref{fig1} (b)) to process the fine-grained embeddings, which enhances the ability to perceive forgery by capturing the relationships between different modalities based on a single-stage transformer. However, they overlook the disharmony among information from different data sources. Since it is unable to discern between the forged and genuine content via the consistency learning, it may lead to confusion and ambiguity for the fine-grained sub-tasks. 

In contrast, we extract the clues of localized forgery for DGM$^4$ from the perspective of consistency. Inconsistency in multi-modal forgery may exist within and across modalities. The intra-modal inconsistency mainly stems from the specific information contained in different heterologous data, which can uniquely identify their sources \cite{zhao2021learning}. For an image, the manifestation of specific information are artifacts \cite{nguyen2024laa} and the data distribution difference which may come from imaging pipelines \cite{huh2018fighting}, encoding approaches \cite{barni2017aligned} and synthesis models. Compared to images, the forgery of text is not obvious, but the coherence of narrative can still be seen as the basis for determining consistency \cite{pu2023deepfake}. Because the above-mentioned inconsistency within a modality mainly manifests in conflicting information with the background, we summarize it as contextual consistency. Meanwhile, the inconsistency between modalities is mainly reflected in the different meanings expressed by two modal data for the same scene, including emotions, subjects, etc. The information between different modalities is often associated through semantics, so we summarize it as semantic consistency. Constructing and supervising the consistency enhances the ability of distinguishing between the forged and authentic content \cite{yang2022unified}. Using consistency-assisted feature extraction and reasoning enhances the interpretability and reliability . At the same time, to better solve fine-grained forgery tasks, the construction of consistency should not be limited to using a global embedding for contrastive learning \cite{shao2023detecting}. We then propose to employ fine-grained consistency learning for each image patch or text token to enhance the perception ability of local regions.

As shown in Fig. \ref{fig1} (c), we propose a novel framework named Contextual-semantic Consistency Learning (CSCL), which tries to unleash the potential of consistency learning for the DGM$^4$ task. Specifically, contextual and semantic consistency decoders are proposed. The former calculates a consistency matrix based on the continuity of context among fine-grained embeddings within one modality, while the latter constructs a consistency matrix based on the semantic similarity between the fine-grained embeddings of one modality and the global embedding of other modalities. 
A consistency loss is introduced to supervise the consistency matrix. After the aforementioned consistency construction,
a forgery-aware reasoning or aggregating module is adopted under the guidance of consistency,
which deeply captures forgery cues and uses the attention mechanism on selective embeddings to alleviate the influence caused by redundant or confused content.
Extensive experiments on the DGM$^4$ datasets \cite{shao2023detecting} show that CSCL can achieve new state-of-the-art results, especially for grounding manipulated content. Ablation study also proves the effectiveness of each proposed modules. Our contributions are summarized as:
\begin{itemize}
    \item We introduce a novel framework named CSCL for the DGM$^4$ task, which focuses on making fine-grained consistency learning and locating manipulated content.
    \item We propose contextual and semantic consistency decoders which seek consistency within and across modalities, respectively. Forgery-aware reasoning and aggregating modules are also used to deeply capture forgery cues.
    \item We confirm the efficacy of CSCL by achieving the state-of-the-art results on DGM$^4$ datasets and greatly improve the accuracy of grounding manipulated content.
\end{itemize}

\section{Related work}
\label{sec:formatting}

\subsection{Face deepfake detection}
In order to ensure security and privacy, many face deepfake detection methods are proposed which could be roughly divided into frequency-based \cite{li2021frequency, woo2022add} and spatial-based methods. Frequency-based methods transform the time domain information of an image into the frequency domain \cite{pei2024deepfake} and conduct further process on the 
transformed feature map. For instance, F$^3$-Net \cite{qian2020thinking} uses a dual-branch structure to explore the artifacts of suspicious images via frequency-aware decomposition and local frequency statistic. HFI-Net \cite{miao2022hierarchical} extracts multi-level frequency-related forgery clues by Global-Local Interaction modules. For spatial-based methods, some works use detail differences as the judgment criteria, including saturation \cite{mccloskey2019detecting}, color \cite{he2019detection}, gradient \cite{tan2023learning}, \textit{etc}. They explore the disharmony \cite{ba2024exposing} and inconsistency between different regions through these details. Another popular classification of spatial-based methods is based on noise \cite{zhou2017two, nguyen2019capsule}, which could be used to identify the local or global differences. For example, NoiseDF \cite{wang2023noise} proposes an efficient
Multi-Head Relative Interaction with depth-wise separable convolutions to detect the underlying noise traces in the deepfake videos.

\subsection{Multi-modal deepfake detection}
With the increase of multi-modal forgery data on the Internet, multi-modal deepfake detection receives widespread attention. The multi-modal methods which use visual and textual information could be roughly divided into out-of-context misinformation detection \cite{abdelnabi2022open, luo2021newsclippings, mu2023self} and fake news detection \cite{jin2017multimodal, khattar2019mvae, wang2018eann}. Out-of-context misinformation detection often use the real image as the evidence to measure the confidence level of the text narrative. For instance, CCN \cite{abdelnabi2022open} utilizes consistency checking between image and text to analyze the reliability of the caption. For fake news detection, most previous methods \cite{ying2023bootstrapping} focus on predict the binary classification which determines news authenticity. For instance, MMFN \cite{zhou2023multi} extracts multi-grained features and fuses them for the binary prediction. HAMMER \cite{shao2023detecting} constructs  the first dataset for DGM$^4$. Many methods \cite{shao2024detecting, li2024towards, wang2024exploiting} are proposed to tackle the DGM$^4$ problem. For example, UFAFormer \cite{liu2024unified} introduces a unified framework which adopts additional frequency domain information to detect visual forgery artifacts. However, existing works can not achieve satisfactory results of grounding manipulated content. 
% To this end, we propose CSCL which conduct forgery aware fine-grained modeling under the guidance of contextual and semantic consistency.
To this end, we propose CSCL which conducts contextual and semantic consistency learning among fine-grained embeddings to help deeper reasoning.
\begin{figure*}[t]
  \centering
   \includegraphics[scale=0.52]{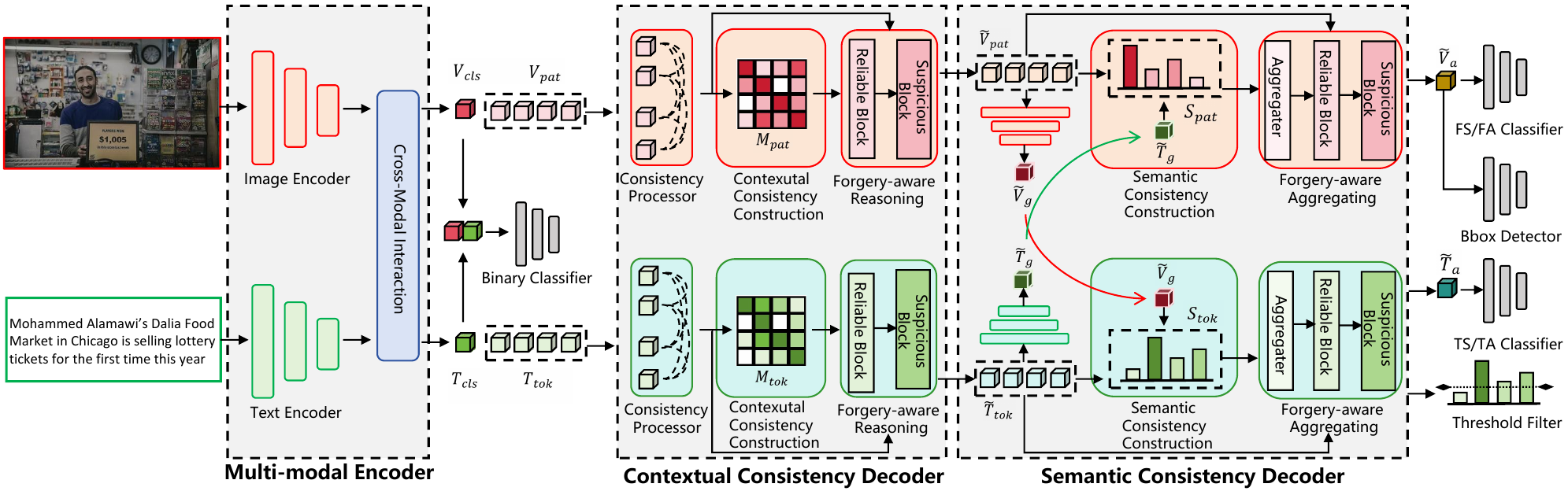}
   \caption{\textbf{The overall architecture of CSCL.} CSCL can be divided into contextual consistency decoder and semantic consistency decoder. These decoders construct fine-grained consistency matrices and a use consistency loss for supervision. In each decoder, a forgery-aware reasoning or aggregating module is used to reduce the interference of confused content and deeply explore forgery cues.
   }
   \label{fig2}
\end{figure*}

\subsection{Consistency learning}

Consistency is a widely used criterion for deepfake detection. Some works measure the intra-modality consistency which calculates similarity scores among feature embeddings \cite{mayer2020exposing, mayer2019forensic}. For instance, Zhou \textit{et al.} \cite{zhou2017two} propose a two-stream
network to estimate the  tampered faces and low-level inconsistency. PCL \cite{zhao2021learning} proposes an end-to-end learning pipeline that measures the image self-consistency with one forward pass. Some works \cite{yang2023masked, yin2023dynamic} find the inter-frame consistency for forgery detection, which calculates the similarity among adjacent frames. For instance, snippet \cite{gu2022delving} conducts local temporal inconsistency learning based on densely sampling adjacent frames. Some works \cite{cozzolino2023audio, oorloff2024avff} measure the inter-modality consistency which mainly estimates the semantic similarity among different modalities. For instance, HAMMER \cite{shao2023detecting} uses the contrastive learning to help the uni-modal encoders better exploit the semantic correlation between image and text. 
However, previous consistency learning can not effectively capture detailed forgery information. In view of this, we propose CSCL which has the following traits: (1) fine-grained consistency, (2) intra-model and inter-modal consistency, and (3) the guidance of consistency to capture forgery cues.
\section{Contextual-semantic consistency learning}

\subsection{Overview}
\label{sec:3.1}

The overall architecture of CSCL is shown in Fig. \ref{fig2}. It is composed of multi-modal encoder, contextual consistency decoder (Section \ref{sec:3.2}) and semantic consistency decoder (Section \ref{sec:3.3}).
Multi-modal encoder extracts uni-modal features and learns correlation between them, while contextual and semantic consistency decoders enhance the ability of model to distinguish and perceive counterfeit content.
We supervise the network by calculating the sub-task loss and the consistency loss (Section \ref{sec:3.4}).

\noindent \textbf{Multi-modal encoder.} The Multi-modal encoder consists of uni-modal (image and text) encoders and cross-modal interaction.  Following previous methods \cite{wang2024exploiting}, we use ViT-B/16 \cite{dosovitskiy2020image} and RoBERTa \cite{liu2019roberta} as the image encoder and text encoder, respectively.
Given the image-text pair, we first divide an image into $n$ patches and insert an image class token. Then, the image encoder encodes them into a sequence of image embeddings. For text inputs, the text encoder is used to process text tokens and inserted text class token into text embeddings.
There may be some distinctive information that differs between the outputs of uni-modal encoders, which is the key clue to distinguishing authenticity. To obtain deeper correlations and find the differences, we use a cross-modal interaction module to produce cross-modal representations. The cross-modal interaction module consists of multiple co-attention layers \cite{dou2022empirical}. In each layer, text and visual features are fed into different transformer blocks independently, and cross-attention are used for feature interaction. The outputs of cross-modal interaction module could be divided into class embeddings ($V_{cls}$ and $T_{cls}$) and fine-grained embeddings ($V_{pat}$ and $T_{tok}$).

% \noindent \textbf{Consistency learning module.}
However, it is difficult to effectively distinguish between genuine and forged content with only an attention mechanism \cite{vaswani2017attention}, so directly using the outputs of multi-modal encoder to predict fine-grained sub-tasks is insufficient. To make the network have the ability of perceiving disharmony between heterogeneous information, we explore the consistency learning among fine-grained image patches (or data tokens) from both semantic and contextual perspectives.
The fine-grained embeddings of the multi-modal encoder's outputs enter contextual consistency decoder and semantic consistency decoder sequentially to conduct deeper feature extraction based on the consistency.
The outputs of the latter decoder are the global aggregated features ($\widetilde{V}_{a}$ and $\widetilde{T}_{a}$) which will be used for further prediction.

\subsection{Contextual consistency decoder}
\label{sec:3.2}
The forged and genuine content comes from different data sources, which often leads to inconsistency between contexts.
Finding this inconsistency is beneficial for accurately locating manipulated regions. In this view, as shown in Fig. \ref{fig2}, we propose contextual consistency decoder to construct intra-modal correlations and extract the forgery clues.

\noindent\textbf{Consistency processor.} Inconsistent context may occur between distant content, so
establishing long-range dependencies is crucial for contextual consistency construction. To this end, we use consistency processor to learn the association among every fine-grained embeddings. It is composed of three standard self-attention layers with position embeddings. We adopt sine-cosine functions followed with MLP layers to calculate the position embeddings.

\begin{figure*}[t]
  \centering
   \includegraphics[scale=0.53]{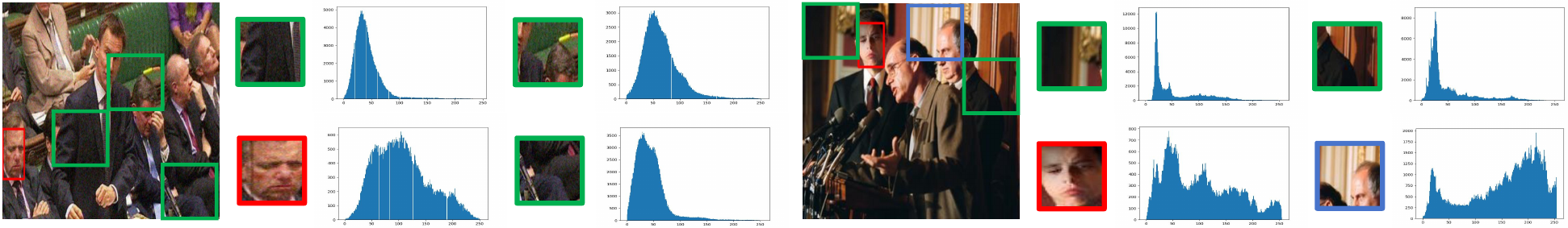}
   \caption{\textbf{ Histogram of genuine, forged and confused image patches.} Genuine patches are marked by \textcolor{green}{green} , forged patches are marked by \textcolor{red}{red} (down right), and confused patch is marked by \textcolor{blue}{blue} (down left). Observe the high-frequency range to determine consistency.
   }
   \label{fig8}
\end{figure*}

\noindent\textbf{Contextual consistency construction.} Through the construction and the supervision of contextual consistency, our model can enhance the distinctiveness of features corresponding to different data sources. We the take image modality as an example to introduce the process of consistency construction. Images contains content-independent or spatially-local information that can uniquely identify their sources \cite{zhao2021learning}. These information may originate from the differences in data distribution between nature and synthetic images. For example, as shown in Fig. \ref{fig8}, we visualize the histogram of genuine and forged images patches, which reflects the tone of localized regions. There are significant differences between the histogram of the forged patches and the real patch in both near and far distance.
Given the outputs of consistency processor $\overline{V}_{pat}=\{\overline{V}_1, ..., \overline{V}_n\}$, for each patch embedding, we
compare it against all the rest to measure their feature consistency, thus obtain a 2D consistency matrix $M_{pat}$ in range of [0, 1], whose size is n$\times$n. Here, $n$ is the number of image patches. To be specific, for a certain embedding pair $\overline{V}_i$ and $\overline{V}_j$,
we calculate the consistency score by $M_{pat}^{(i, j)}$ Eq. \ref{eq2}.
\begin{equation}
\label{eq2}
M_{pat}^{(i, j)} = \frac{1}{2}(\frac{\varphi(\overline{V}_i)^T\varphi(\overline{V}_j)}{|\varphi(\overline{V}_i)|.|\varphi(\overline{V}_j)|}+1)
\end{equation}
where $\varphi(.)$ is the multi-layer perception (MLP) function, and $|.|$ denotes the 2-norm of the embedding. We add 1 to the cosine similarity and then divide by 2 to scale the consistency score between 0 and 1.

For the ground truth of the image consistency matrix $\overline{M}_{pat}$, 
if the patches corresponding to an item in the matrix are both manipulated or both not, it is set to `1', which means they come from the same data source; otherwise, it is set to `0'. Similarly, we could obtain the consistency matrix of text $M_{tok}$ and its corresponding ground truth $\overline{M}_{tok}$. The supervision process will be detailed in Section \ref{sec:3.4}. 

\noindent\textbf{Forgery-aware reasoning.} 
Models may encounter confusion when determining the consistency of certain content, mainly due to the insignificant features of these content or puzzling pattern. For example, as shown in Fig. \ref{fig8}, the histogram of blue box is neither similar to genuine nor forged patches. In this view, we conduct additional forgery-aware reasoning which learns correlation on selective embeddings to reduce the attention to confused content.
Using a certain image embedding $\overline{V}_{pat}^{i}$ as the example, we select $k$ most similar features as reliable content $\overline{V}_{pat}^{r}$ and $k$ most unsimilar features as suspicious content $\overline{V}_{pat}^{s}$ from $\overline{V}_{pat}$ based on contextual consistency matrix $M_{pat}$. Then, the reliable block is used to model the correlation between  $\overline{V}_i$ and $\overline{V}_{pat}^{r}$, and the suspicious block is used to model the correlation between $\overline{V}_i$ and  $\overline{V}_{pat}^{s}$. Both reliable and suspicious blocks are composed of attention mechanism \cite{vaswani2017attention} and residual connection \cite{he2016deep}. We process each patch and token embeddings in the same way, and obtain $\widetilde{V}_{pat}$ and $\widetilde{T}_{tok}$.

\subsection{Semantic consistency decoder}
\label{sec:3.3}
There may be semantic inconsistency between text and image. For example, the genuine image depicts a joyful scene, while the forged text contains negative words, which can serve as a basis for forgery detection. To this end, as shown in Fig. \ref{fig2}, we propose semantic consistency decoder which constructs correlation between image and text.

\noindent \textbf{Semantic consistency construction.}
Since local content lacks enough semantics and the content of another modality may be partially forged, 
it is difficult to achieve effective supervision to the consistency between each image patch and each text token. 
To solve this issue, we aggregate the fine-grained embeddings from another modality into a global embedding and calculate the similarity with it. Using the construction of image matrix as the example,
We first calculate global embedding of text $\widetilde{T}_{g}$ via Eq. \ref{eq5}.
\begin{equation}
\label{eq5}
\widetilde{T}_{g} = \Phi_{t}(\sigma_t(t, \widetilde{T}_{tok}, \widetilde{T}_{tok})),
\end{equation}
where $t$ is the randomly initialized embedding used to represent the entire sentence. $\Phi_{i}(.)$ and $\Phi_{t}(.)$ are the MLP functions. $\sigma_t(.)$ is the attention functions. For each patch embedding, we compare it with $\widetilde{T}_{g}$, and obtain semantic consistency matrix of image $S_{pat}$, whose size is n$\times$1. For a certain patch embedding $\widetilde{V}_{pat}^i$, the consistency sore $S_{pat}^{(i)}$ is calculated by Eq. \ref{eq6}.
\begin{equation}
\label{eq6}
S_{pat}^{(i)} = \frac{1}{2}(\frac{\varphi(\widetilde{V}_{pat}^i)^T\varphi(\widetilde{T}_{g})}{|\varphi(\widetilde{V}_{pat}^i)|.|\varphi(\widetilde{T}_{g})|}+1).
\end{equation}
For the ground truth of consistency matrix $\overline{S}_{pat}$, if the corresponding content is not under manipulation, it is set to `1', otherwise set to ‘0’. Similarly, we could obtain the semantic consistency matrix $S_{tok}$ of text and the ground truth $\overline{S}_{tok}$.

\noindent\textbf{Forgery-aware aggregating.}
% To fully unleash the potential of semantic consistency learning for forgery detection,
To deeply capture forgery cues and reduce confusion, the semantic consistency matrix is used to give the guidance for further process.
% We obtain the most reliable and suspicious content and use forgery-aware reasoning module to calculate the aggregated embedding for later prediction.
Using a image branch as the example, we adopt forgery-aware aggregating to extract aggregated embedding $\widetilde{V}_{a}$ by Eq. \ref{eq8}.
\begin{equation}
\label{eq8}
\widetilde{V}_{a} = f_a(\sigma_i(x, \widetilde{V}_{pat}, \widetilde{V}_{pat}),\widetilde{V}_{pat}^{r}, \widetilde{V}_{pat}^{s}),
\end{equation}
where $\sigma_i(.)$ is the attention function, $f_a(.)$ is the forgery-aware reasoning (as mentioned in Section \ref{sec:3.2}). $\widetilde{V}_{pat}^{r}$ and $\widetilde{V}_{pat}^{s}$ are the $k$ most reliable and suspicious patch embeddings, respectively. $x$ is a randomly initialized embedding that represents the entire image. The aggregated embedding of text $\widetilde{T}_{a}$ is calculated in the same way.

\noindent \textbf{Threshold Filter.} For grounding text manipulation, threshold filter is used to make the decision based on the consistency score $S_{tok}$. This means that we no longer need to provide additional prediction head and supervision for grounding text manipulation as previous methods \cite{wang2024exploiting,shao2023detecting}.
The reason is that the main evidence for determining the authenticity of text is its similarity to image, using consistency scores for decision can more explicitly represent this process and achieve more flexible results. In addition, we experimentally prove this viewpoint in Section \ref{sec:4.4}.

\subsection{Prediction and loss}
\label{sec:3.4}

For prediction, The class embeddings ($V_{cls}$ and $T_{cls}$) are concatenated and inputted into the binary classifier. Image aggregated feature $\widetilde{V}_{a}$ is used to predict the fake face bounding box and the face fine-grained type, including face swap (FS) and face attributes (FA) manipulations. Text aggregated feature $\widetilde{T}_{a}$ is used to predict the text fine-grained type, including text swap (TS) and text attributes (TA) manipulations. Different from previous methods \cite{shao2023detecting,shao2024detecting} which use 
token embeddings to predict whether the word is replaced, we adopt consistency scores between each token and the image as the criteria. All the used classifiers or decoders are composed of MLP.

For supervision, we first introduce consistency loss. Using contextual consistency matrix as the example, given image matrix $M_{pat}$, text matrix $M_{tok}$ and their ground truth $\overline{M}_{pat}$ and $\overline{M}_{tok}$.
The loss $L_m$ can be obtained by Eq. \ref{eq3}.
\begin{equation}
\label{eq3}
\footnotesize
\begin{aligned}
L_c = \frac{1}{n^2}\sum_{i=1}^{n^2}(\overline{M}_{pat}^{(i)}log(M_{pat}^{(i)})&+(1-\overline{M}_{pat}^{(i)})log(1-M_{pat}^{(i)})), \\
+ \frac{1}{m^2}\sum_{j=1}^{m^2}(\overline{M}_{tok}^{(j)}log(M_{tok}^{(j)})&+(1-\overline{M}_{tok}^{(j)})log(1-M_{tok}^{(j)})), \\
% L_{c} = L_c^{p} &+ L_c^{t},
\end{aligned}
\end{equation}
where $n$ and $m$ are the side length of image and text matrices, respectively.
Similarly, we could obtain the loss $L_s$ of semantic consistency matrix. For other sub-tasks, we use the same supervision function following \cite{wang2024exploiting}. 
\section{Experiments}

\subsection{Dataset and metrics}
The experiments are conducted on the DGM$^4$ \cite{shao2023detecting} dataset which contains 230 image-text news pairs, including 77426 genuine pairs and 152574 manipulated pairs. The real-world news source of DGM$^4$ includes The Guardian, BBC, USA TODAY, and The Washington Post. There are totally four types of manipulation, including face swap (FS), face attribute (FA), text swap (TS), and text attribute (TA). Following previous methods \cite{shao2023detecting, liu2024unified, wang2024exploiting}, we use accuracy (ACC), area under the receiver operating characteristic
curve (AUC), and equal error rate (EER) as the metrics for binary classification. We evaluate the results of fine-grained
classification through mean average precision (MAP), average per-class F1 (CF1), and overall F1 (OF1). For manipulated image grounding, mean intersection over union (IoU$_m$), the IoU at thresholds of 0.5 (IoU50) and 0.75 (IoU75) are used for evaluation. We evaluate manipulated text grounding results via precision, recall, and F1 score.

\subsection{Implement details}
The size of images is set to 256$\times$256, while the length of text is padded to 50. Following Wang et al. \cite{wang2024exploiting}, we use the ViT-B/16 \cite{dosovitskiy2020image} as the image encoder and RoBERTa \cite{liu2019roberta} as the text encoder, and the pre-trained weights of backbones are loaded from METER \cite{dou2022empirical}. The number of co-attention layers is set to 6. The number of attention layers in consistency processor is set to 3. 
The AdamW \cite{loshchilov2017fixing} is used as the optimizer with a weight decay of 0.02, and the learning rate is set to $1\times10^{-5}$.
We train CSCL with 50 epochs on 8 A100 GPUs, the batch size is set to 32 on each GPU.

\subsection{Comparison with the state-of-the-art methods}

\begin{table*}[!t] 
\renewcommand{\arraystretch}{1.10} 
    \setlength{\tabcolsep}{1mm}
    \begin{center}
    \caption{\textbf{Comparison of state-of-the-art methods for DGM$^4$.} $\downarrow$ means less is better. The best results is bold. PR. represents 
    precision, while RE. represents recall.}
    \label{tab1}
    \resizebox{\textwidth}{!}{  
    \begin{tabular}
    %{cc|ccc|ccc|ccc|ccc} 
    {
    p{15pt}<{\raggedright}
    p{78pt}<{\raggedright} p{50pt}<{\raggedright} 
    p{32pt}<{\centering}p{32pt}<{\centering}p{32pt}<{\centering}
    p{32pt}<{\centering}p{32pt}<{\centering}p{32pt}<{\centering}
    p{32pt}<{\centering}p{32pt}<{\centering}p{32pt}<{\centering} 
    p{32pt}<{\centering}p{32pt}<{\centering}p{32pt}<{\centering} 
    }   
    \hline
    & \multirow{2}{*}{Method} & \multirow{2}{*}{Ref.}
    &\multicolumn{3}{c}{Binary Cls}&\multicolumn{3}{c}{Multi-label Cls}&\multicolumn{3}{c}{Image Grounding}&\multicolumn{3}{c}{Text Grounding}\\  
    & & &AUC&EER$\downarrow$&ACC&mAP&CF1&OF1&IoU$_m$&IoU50&IoU75&PR.&RE.&F1 \\
    \hline
    \multirow{7}{*}{\rotatebox{90}{Img Sub.}} & TS \cite{luo2021generalizing}&CVPR'21&$91.80$&$17.11$&$82.89$&-&-&-&$72.85$&$79.12$&$74.06$&-&-&-\\
    & MAT \cite{zhao2021multi}&CVPR'21&$91.31$&$17.65$&$82.36$&-&-&-&$72.88$&$78.98$&$74.70$&-&-&-\\
    & HAMMER \cite{shao2023detecting}&CVPR23&$94.40$&$13.18$&$86.80$&-&-&-&$75.69$&$82.93$&$75.65$&-&-&-\\
    & HAMMER++ \cite{shao2024detecting}&TPAMI'24&$94.69$&$13.04$&$86.82$&-&-&-&$75.96$&$83.32$&$75.80$&-&-&-\\
    & ViKI \cite{li2024towards}&IF'24&$91.85$&$15.92$&$84.90$&-&-&-&$75.93$&$82.16$&$74.57$&-&-&-\\
    & UFAFormer \cite{liu2024unified}&IJCV'24&$94.88$&$12.35$&$87.16$&-&-&-&$77.28$&$85.46$&$78.29$&-&-&-\\
    \rowcolor{Gray} & Ours&CVPR'25&\textbf{97.15}&\textbf{8.81}&\textbf{91.18}&-&-&-
    &\textbf{82.78}&\textbf{90.19}&\textbf{86.31}&-&-&-\\
    \hline
    \multirow{7}{*}{\rotatebox{90}{Text Sub.}} & BETR \cite{kenton2019bert}&NAACL'19&$80.82$&$28.02$&$68.98$&-&-&-&-&-&-&$41.39$&$63.85$&$50.23$\\
    & LUKE \cite{yamada2020luke}&EMNLP'20&$81.39$&$27.88$&$76.18$&-&-&-&-&-&-&$50.52$&$37.93$&$43.33$\\
    & HAMMER \cite{shao2023detecting}&CVPR'23&$93.44$&$13.83$&$87.39$&-&-&-&-&-&-&$70.90$&$73.30$&$72.08$\\
    & HAMMER++ \cite{shao2024detecting}&TPAMI'24&$93.49$&$13.58$&$87.81$&-&-&-&-&-&-&$72.70$&$72.57$&$72.64$\\
    & ViKI \cite{li2024towards}&IF'24&$92.31$&$15.27$&$85.35$&-&-&-&-&-&-&$78.46$&$65.09$&$71.15$\\
    & UFAFormer \cite{liu2024unified}&IJCV'24&$94.11$&$12.61$&$84.71$&-&-&-&-&-&-&$81.13$&$70.73$&$75.58$\\
    \rowcolor{Gray} & Ours&CVPR'25&\textbf{96.38}&\textbf{9.53}&\textbf{89.74}&-&-&-&-&-&-&\textbf{82.88}&\textbf{77.92}&\textbf{80.32}\\
    \hline
    \multirow{8}{*}{\rotatebox{90}{Entire Dataset}} & CLIP \cite{radford2021learning}&ICML'21&$83.22$&$24.61$&$76.40$&$66.00$&$59.52$&$62.31$&$49.51$&$50.03$&$38.79$&$58.12$&$22.11$&$32.03$\\
    & ViLT  \cite{kim2021vilt}&ICML'21&$85.16$&$22.88$&$78.38$&$72.37$&$66.14$&$66.00$&$59.32$&$65.18$&$48.10$&$66.48$&$49.88$&$57.00$\\
    & HAMMER \cite{shao2023detecting}&CVPR'23&$93.19$&$14.10$&$86.39$&$86.22$&$79.37$&$80.37$&$76.45$&$83.75$&$76.06$&$75.01$&$68.02$&$71.35$\\
    & HAMMER++ \cite{shao2024detecting}&TPAMI'24&$93.33$&$14.06$&$86.66$&$86.41$&$79.73$&$80.71$&$76.46$&$83.77$&$76.03$&$73.05$&$72.14$&$72.59$\\
    & ViKI \cite{li2024towards}&IF'24&$93.51$&$13.87$&$86.67$&$86.58$&$81.07$&$80.10$&$76.51$&$83.95$&$75.77$&\textbf{77.79}&$66.06$&$73.44$\\
    & UFAFormer \cite{liu2024unified}&IJCV'24&$93.81$&$13.60$&$86.80$&$87.85$&$80.31$&$81.48$&$78.33$&$85.39$&$79.20$&$73.35$&$70.73$&$72.02$\\
    & Wang et al. \cite{wang2024exploiting}&ICASSP'24&$95.11$&$11.36$&$88.75$&$91.42$&$83.60$&$84.38$&$80.83$&$88.35$&$80.39$&$76.51$&$70.61$&$73.44$\\
    \rowcolor{Gray} & Ours&CVPR'25&\textbf{96.34}&\textbf{9.88}&\textbf{90.32}&\textbf{92.48}&\textbf{86.19}&\textbf{86.92}&\textbf{84.07}&\textbf{90.48}&\textbf{87.17}&$75.33$&\textbf{77.95}&\textbf{76.62}\\ \hline
    \end{tabular}
    }
\end{center}
\vspace{-0.5cm}
\end{table*}

As shown in Table \ref{tab1}, we compare our proposed CSCL with SOTA uni-modal and multi-modal frameworks. For multi-modal methods, we beat all the existing methods by achieving 96.34\% AUC, 92.48\% mAP, 84.07\% IoU$_m$ and 76.62\% F1 on binary classification, multi-label classification, image grounding and text grounding, respectively. What's more, significant improvements are achieved on grounding image and text manipulations. 
\begin{figure}[t]
  \centering
   \includegraphics[width=6.8cm, height=2.5cm]{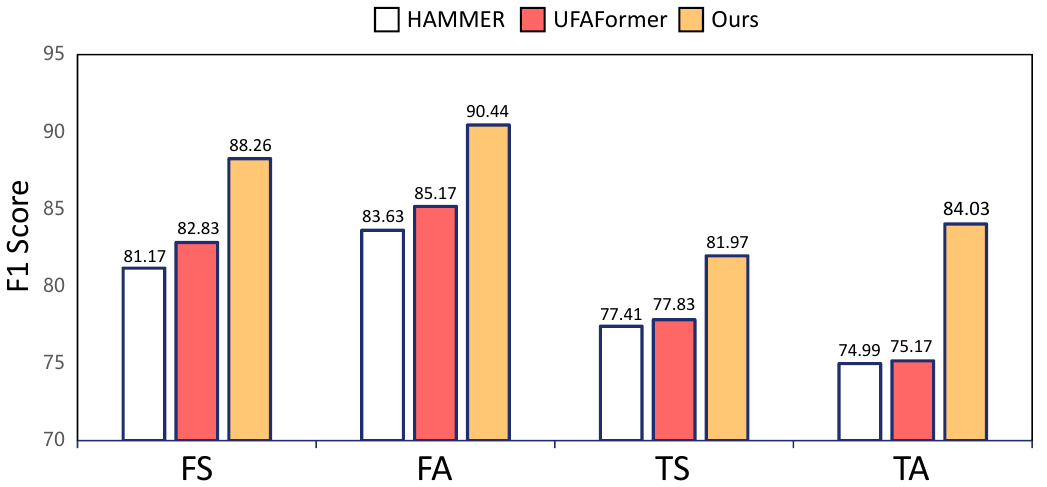}
   \caption{\textbf{F1 scores of four manipulation types
    in fine-grained manipulation classification.} FS, FA, TS, TA denotes face swap, face attribute, text swap, text attribute, respectively. 
   }
   \label{fig6}
   \vspace{-0.6cm}
\end{figure}
Specifically, compared to recently proposed Wang et al. \cite{wang2024exploiting}, CSCL gains +3.24\%, +2.13\%, +6.78\% and +3.18\% on IoU$_m$, IoU50, IoU75 and F1, respectively. It should be noted that precision and recall are two mutual inhibition metrics on text grounding. When measuring its performance, we often consider the level of the comprehensive indicator F1. For uni-modal methods, following previous methods \cite{shao2023detecting, liu2024unified}, we divide the entire dataset into two single-modal forgery sub-datasets. CSCL surpasses all the methods on both image and text sub-datasets. For instance, CSCL exceeds UFAFormer \cite{liu2024unified} by a large margin of +5.50\% IoU$_m$ and +4.74\% F1 on the image and text sub-datasets, respectively. The above results demonstrate that our method significantly improves forgery localization under different types of forgery. Besides, as shown in Fig. \ref{fig6}, we visualize the F1 score of four different manipulation types in fine-grained manipulation type classification. It could be observed that CSCL significantly surpasses UFAFormer \cite{liu2024unified} in all manipulation types, especially +5.43\% on face swap and +8.86\% on text attribute.

\subsection{Ablation study}
\label{sec:4.4}
We first give a brief description of our baseline. We remove contextual and semantic consistency decoders of CSCL and directly use the outputs of cross-modal interaction for the later prediction. For image localization, the LPAA \cite{he2019detection} module is used to aggregate fine-grained embeddings into global embedding, which is inputted to Bbox detector. For grounding text manipulation, we use token embeddings to predict whether the word is replaced.
\begin{figure*}[t]
  \centering
   \includegraphics[scale=0.48]{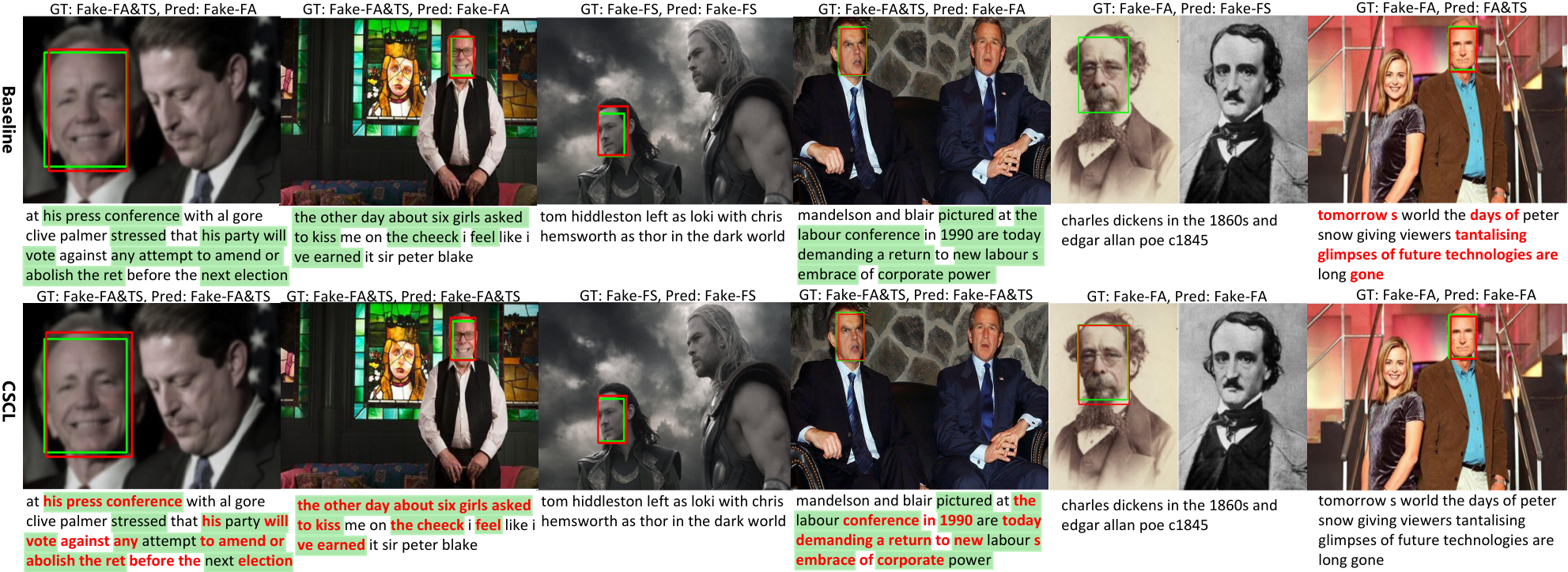}
   \caption{\textbf{Visualization of detection and grounding results.} Here, \textcolor{red}{red} box and text indicate the prediction of manipulated faces and words, while \textcolor{green}{green} box and text represent the corresponding ground truth.
   }
   \label{fig3}
\end{figure*}
% \begin{table*}
\begin{table*}
\renewcommand{\arraystretch}{1.10}
    \centering
    \caption{\textbf{Ablation of different kinds of consistency decoder in the DGM$^4$ dataset.} C.I., C.T., S.I. and S.T. denote using contextual consistency decoder on image, using contextual consistency decoder on text, using semantic consistency decoder on image and using semantic consistency decoder on text, respectively. When C.I., C.T., S.I. and S.T. are not used at the same time, it means the baseline.} 
    \label{tab2}
    \setlength{\tabcolsep}{1.7mm}
    \resizebox{\textwidth}{!}{  
    \begin{tabular}
    % {cccc|ccc|ccc|ccc|ccc}
    {
    p{15pt}<{\raggedright}p{15pt}<{\raggedright}p{15pt}<{\raggedright}p{15pt}<{\raggedright}
    p{32pt}<{\centering}p{32pt}<{\centering}p{32pt}<{\centering}
    p{32pt}<{\centering}p{32pt}<{\centering}p{32pt}<{\centering}
    p{32pt}<{\centering}p{32pt}<{\centering}p{32pt}<{\centering} 
    p{32pt}<{\centering}p{32pt}<{\centering}p{32pt}<{\centering}
    % p{20pt}<{\centering}  
    }  
    \hline
    \multicolumn{4}{c}{Components}&\multicolumn{3}{c}{Binary Cls}&\multicolumn{3}{c}{Multi-label Cls}&\multicolumn{3}{c}{Image Grounding}&\multicolumn{3}{c}{Text Grounding} \\
    C.I.&C.T.&S.I.&S.T.&AUC&EER$\downarrow$&ACC&mAP&CF1&OF1&IoU$_m$&IoU50&IoU75&PR.&RE.&F1\\
    \hline
    &&&&$96.02$&$10.24$&$89.97$&$91.97$&$85.36$&$86.01$&$81.21$&$88.88$&$80.26$&\textbf{79.18}&$69.09$&$73.79$\\
    % \hline
    \checkmark&\checkmark&&&$96.23$&$10.02$&$90.14$&$92.38$&$85.96$&$86.68$&$83.70$&$90.19$&$86.94$&$79.10$&$72.23$&$75.51$\\
    &&\checkmark&\checkmark&$96.17$&$9.98$&$90.22$&$92.28$&$86.02$&$86.75$&$81.60$&$88.92$&$82.87$&$75.55$&$76.68$&$76.10$\\
    \checkmark&&\checkmark&&$96.22$&$9.93$&$90.12$&$92.25$&$85.99$&$86.70$&$83.92$&$90.43$&$86.99$&$78.56$&$70.39$&$74.25$\\
    &\checkmark&&\checkmark&$96.15$&$10.06$&$90.09$&$92.20$&$86.16$&$86.88$&$81.06$&$88.88$&$79.78$&$72.74$&\textbf{79.60}&$76.02$\\
    % \hline
    \rowcolor{Gray} 
\checkmark&\checkmark&\checkmark&\checkmark&\textbf{96.34}&\textbf{9.88}&\textbf{90.32}&\textbf{92.48}&\textbf{86.19}&\textbf{86.92}&\textbf{84.07}&\textbf{90.48}&\textbf{87.17}&$75.33$&$77.95$&\textbf{76.62}\\
    \hline
    \end{tabular}
    }
    % \vspace{-0.5cm} 
\end{table*}
% \begin{table}
\begin{table}
\renewcommand{\arraystretch}{1.10}
    \centering
    \caption{\textbf{Ablation of different backbones in the DGM$^4$ task.} $\bigtriangleup$ denotes the momentum version of ALBEF \cite{shao2023detecting} backbone, $\bigtriangledown$ denotes the normal version of ALBEF backbone and $\diamondsuit$ denotes the METER \cite{dou2022empirical} backbone.
    }
    \label{tab3}
    \setlength{\tabcolsep}{0.8mm}
    \resizebox{0.48\textwidth}{!}{ 
    \begin{tabular}
    % {cc|cc|cccc}
    {
    p{45pt}<{\raggedright} 
    p{40pt}<{\raggedright}p{40pt}<{\raggedright}p{40pt}<{\raggedright}p{40pt}<{\raggedright}p{40pt}<{\raggedright}p{40pt}<{\raggedright}
    % p{20pt}<{\centering}   
    } 
    \hline
    \multirow{2}{*}{Method}&\multicolumn{6}{c}{Metric} \\
    &IoU$_m$&IoU50&IoU75&PR.&RE.&F1\\
    \hline
    Baseline$^\bigtriangleup$&$77.21$&$84.74$&$75.41$&$75.99$&$67.95$&$71.75$\\
    \rowcolor{Gray} CSCL$^\bigtriangleup$&$79.05$&$85.90$&$80.61$&$73.27$&$72.35$&$72.86$\\
    Baseline$^\bigtriangledown$&$77.45$&$84.80$&$76.05$&$76.71$&$63.73$&$69.62$\\
    \rowcolor{Gray} CSCL$^\bigtriangledown$&$79.37$&$86.05$&$80.99$&$72.63$&$71.92$&$72.28$\\
    Baseline$^\diamondsuit$&$81.21$&$88.88$&$80.26$&$79.18$&$69.09$&$73.79$\\
    \rowcolor{Gray} CSCL$^\diamondsuit$&$84.07$&$90.48$&$87.17$&$75.33$&$77.95$&$76.62$\\
    \hline
    \end{tabular}
    }
    \vspace{-0.4cm}
\end{table}
\\\noindent\textbf{Effectiveness of different consistency learning.} As shown in Table \ref{tab2}, we can obtain two conclusions. First, both contextual consistency learning and semantic consistency learning contribute to the results (as shown in Lines 1, 2 and 3). Second, simultaneously using contextual and semantic consistency learning to single modality (image or text) improves the performance (as shown in Lines 1, 4 and 5).
Moreover, compared to detection tasks, CSCL has a more significant improvement in grounding tasks. Using CSCL improves baseline IoU$_m$ and F1 score by 2.86\% and 2.83\%, respectively (as shown in Line 1 and 6).
As shown in Fig. \ref{fig3}, we visualize the detection and grounding results between CSCL and the baseline on the DGM$^4$ dataset.
\begin{figure*}[t]
  \centering
   \includegraphics[scale=0.5]{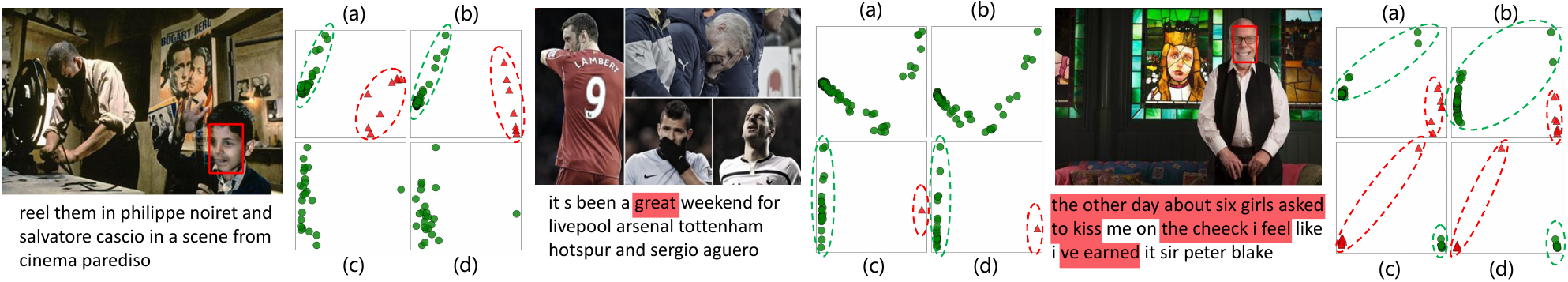}
   \caption{\textbf{Visualization of fine-grained features distribution used in constructing consistency.} (a) contextual consistency of an image, (b) semantic consistency of an image, (c) contextual consistency of text, and (d) semantic consistency of text. The forged content is marked by \textcolor{red}{red}. The \textcolor{green}{green} circle means the features of genuine patches or tokens, while \textcolor{red}{red} triangle means the forged embeddings.
   }
   % \vspace{-0.2cm}
   \label{fig4}
\end{figure*}
\\\noindent\textbf{Effectiveness to different backbones.} As shown in Table \ref{tab3}, we also compare the effectiveness of CSCL on the DGM$^4$ task with other backbones. Experiments show that CSCL can also significantly improve image and text forgery localization results on both normal and momentum version ALBEF \cite{li2021align} backbones.\\
\noindent\textbf{Effectiveness of each components.} As shown in Table \ref{tab5}, we explore the effectiveness of each component in contextual and semantic consistency decoders. Both consistency construction and forgery-aware reasoning (or aggregating) contribute to the performance. The consistency processor in contextual consistency decoder creates enduring connections and improves the comprehension of contextual features. Using consistency scores to select the replaced words in semantic consistency decoders makes the prediction process pay more attention the the semantic correlation and boost the performance. We also explore the most suitable number of image patches and text tokens in forgery-aware reasoning (or aggregating) module. As shown in Fig. \ref{fig7}, the number of image patches should be set to 16, while the number of text tokens should be set to 8. We suppose that this phenomenon is related to the average area/quantity of remarkable content in each modality. Using too few image patches or text tokens may result in insufficient modeling, while using too many may lead to interference of irrelevant regions. As shown in Fig. \ref{fig8}, the F1 score remains relatively stable as the threshold varies from 0.1 to 0.9, demonstrating that CSCL effectively distinguishes between similar and dissimilar content. Finally, we select a threshold of 0.5.

\begin{table*}[!t] 
\renewcommand{\arraystretch}{1.10}
\setlength{\tabcolsep}{1.5mm}
\centering
\caption{\textbf{Ablation study of each component.} $\dag$: we concatenate fine-grained token embeddings and text aggregated feature in channel dimension to predict whether the word is replaced. When using Threshold Filter, the aforementioned process will not be conducted. }
\label{tab5}
\resizebox{\textwidth}{!}{ 
\begin{tabular}
% {ccccc|ccccc}
{
    p{140pt}<{\centering}p{25pt}<{\centering}p{25pt}<{\centering}
    p{25pt}<{\centering}p{25pt}<{\centering}p{25pt}<{\centering}p{25pt}<{\centering}p{140pt}<{\centering}
    p{25pt}<{\centering}p{25pt}<{\centering}p{25pt}<{\centering} 
    p{25pt}<{\centering}p{25pt}<{\centering}p{25pt}<{\centering}
    % p{20pt}<{\centering}  
    }  
    \hline
    \multicolumn{7}{c}{Contextual Consistency Decoder}&\multicolumn{7}{c}{Semantic Consistency Decoder}\\
    Details&IoU$_m$&IoU50&IoU75&PR.&RE.&F1&Details&IoU$_m$&IoU50&IoU75&PR.&RE.&F1\\
    \hline
    Baseline&$81.21$&$88.88$&$80.26$&\textbf{79.18}&$69.09$&$73.79$&Baseline&$81.21$&$88.88$&$80.26$&\textbf{79.18}&$69.09$&$73.79$\\
    +Consistency Processor&$83.37$&$90.06$&$86.16$&$78.03$&$70.94$&$74.31$&+Semantic Consist. Construction&$81.13$&$88.76$&$81.82$&$78.83$&$70.78$&$74.58$\\
    +Contextual Consist. Construction&$83.52$&$89.88$&$86.36$&$78.65$&$71.65$&$74.98$&+Forgery-aware Aggregating $^\dag$ &$81.36$&$88.79$&$82.69$&$78.65$&$72.47$&$75.43$\\
    +Forgery-aware Reasoning  &\textbf{83.70}&\textbf{90.19}&\textbf{86.94}&$79.10$&\textbf{72.23}&\textbf{75.51}&+Threshold Filter&\textbf{81.60}&\textbf{88.92}&\textbf{82.87}&$75.55$&\textbf{76.68}&\textbf{76.10}\\
    \hline
\end{tabular}
}
% \vspace{-0.5cm}
\end{table*}

\begin{figure}[t]
  \centering
   \includegraphics[scale=0.30]{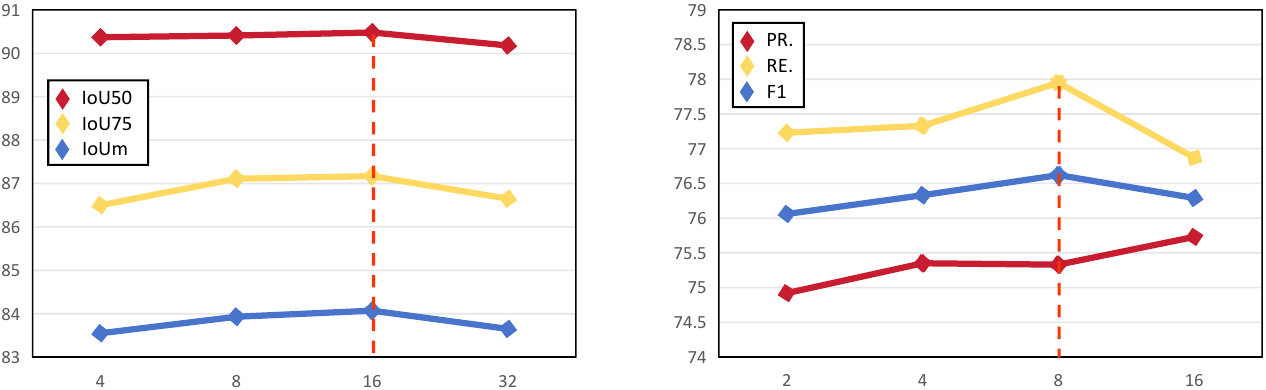}
   \caption{\textbf{The number ablation of image patches (left) and text tokens (right) in forgery-aware reasoning (and aggregating).}
   }
   \vspace{-0.5cm}
   \label{fig7}
\end{figure}

\subsection{Discussion}
\begin{figure}[t]
  \centering
   \includegraphics[width=6.8cm, height=1.6cm]{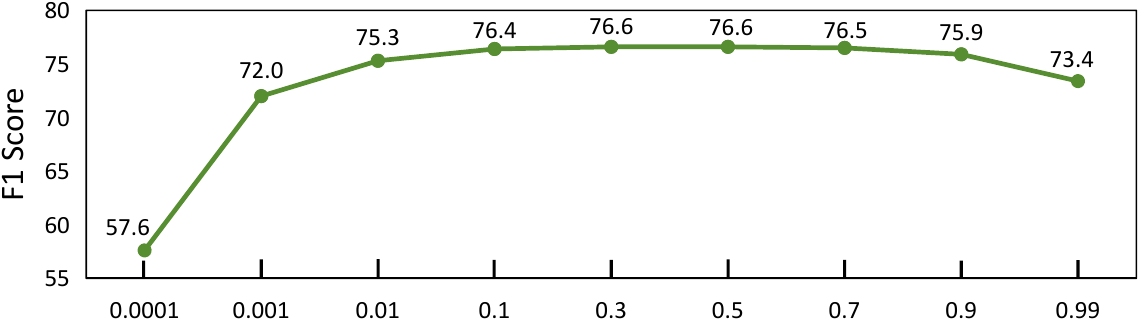}
   \caption{\textbf{Threshold value selection in Threshold Filter.} 
   }
   \label{fig8}
\vspace{-0.5cm}
\end{figure}
The ability of distinguishing features corresponding to different source data is an important prerequisite for implementing CSCL. 
As shown in Fig. \ref{fig4}, we visualize the distribution of features used in consistency construction. We use PCA to compress high-dimensional features into two dimensions for visualization. We totally select three different scenarios for presentation, including manipulating only on images, only on text, and manipulating on both.
It could be noticed that the interface between genuine and forged features can be found in different types of manipulation, and the features of the same type tend to cluster within a region.

\section{Conclusion}
In this paper, we propose a framework named CSCL to make consistency learning and increase the performance of the DGM$^4$ task. Specifically, it consists of contextual and semantic consistency decoders. In each consistency decoder, a consistency matrix is first constructed, and then forgery-aware reasoning or aggregating is conducted under the guidance of consistency. The proposed CSCL can effectively increase the distinctness between forged and genuine content and also find localized forgery clues. Extensive experiments and visualizations demonstrate the effectiveness of our method, especially for grounding manipulation. 
\clearpage
\section*{Acknowledgments}
This work was supported in part by Chinese National Natural Science Foundation Projects U23B2054, 62276254, 62306313, 62206276, the Beijing Science and Technology Plan Project Z231100005923033, Beijing Natural Science Foundation L221013, and InnoHK program.

{
    \small
    \bibliographystyle{ieeenat_fullname}
    \bibliography{main}
}

% WARNING: do not forget to delete the supplementary pages from your submission 
% \input{sec/X_suppl}

\end{document}